\documentclass[runningheads]{llncs}

\usepackage[T1]{fontenc}
\usepackage{graphicx}
\usepackage{amsmath,amssymb}
\usepackage{booktabs}
\usepackage{multirow}
\usepackage{makecell}
\usepackage{xcolor}
\usepackage{subcaption}
\usepackage{xspace}
\usepackage{placeins} 
\usepackage{float}    
\usepackage{url}
\usepackage[hidelinks]{hyperref}


\newcommand{\ourmethod}{\textbf{AnatoProto}\xspace}

\begin{document}

\title{Anatomy-Guided Foundation Model Adaptation with Within-Case Prototype Supervision for Standard Plane Detection in Fetal Ultrasound Blind Sweeps}

\titlerunning{Anatomy-Guided FM Adaptation for Standard Plane Detection}
\author{Yuzhe Zhao}
\authorrunning{Y. Zhao}
\institute{School of Computing, University of Leeds, Leeds, UK\\
\email{clawspeaker@gmail.com}}

\maketitle

\begin{abstract}
Detecting the fetal abdominal circumference standard plane in low-cost obstetric \emph{blind sweeps} is a highly imbalanced frame-classification problem in which positive frames account for less than 3\% of a sequence, form short contiguous segments, and are poorly handled by off-the-shelf ultrasound and vision foundation models. We propose \ourmethod, a lightweight sequence-level framework that adapts a frozen BiomedCLIP encoder to fetal blind sweeps through four complementary components: (i) an \emph{anatomy-weighted spatial pooling} that uses nnU-Net abdominal-region probabilities as a low-dimensional \emph{spatial} prior to reweight BiomedCLIP patch tokens, so that the frozen \emph{semantic} features are aggregated onto anatomically meaningful regions; (ii) a \emph{within-case prototype loss} that pulls each frame embedding toward the mean of the positive frames of the same sweep, exploiting case-level structure that is unavailable at the frame level; (iii) a three-stage \emph{cascade refinement} (frame$\rightarrow$segment$\rightarrow$case-level rejecter) that lifts the prediction unit from noisy frames to structurally-constrained segments; and (iv) a \emph{hybrid prediction head} that jointly models per-frame stability and inter-frame boundary transitions to suppress boundary false positives. On the ACOUSLIC-AI benchmark, \ourmethod reaches a test F1 of \textbf{67.72}, outperforming the strongest external foundation-model baseline (FetalCLIP + PRS, F1 = 54.52) by \textbf{+13.20} F1 and the strongest video temporal-action-detection baseline (TriDet + PRS) by \textbf{+15.76} F1. A dedicated synergy study, backed by pre- and post-training embedding geometry and paired-bootstrap confidence intervals, shows that the prototype loss and the anatomy-weighted pooling are not additive: applied alone the prototype loss significantly reduces recall by $12$ points, but combined with the anatomy-weighted pooling the same loss significantly \emph{increases} recall by $6.5$ points -- a sign-flip that we trace to the accuracy of the within-case prototype, using recall and embedding-geometry evidence rather than F1 alone.

\end{abstract}

\section{Introduction}
Accurate measurement of the fetal abdominal circumference (AC) at the standard plane is one of the most important biometric indicators of intrauterine growth restriction and low birth weight, yet in low- and middle-income settings the sonographic expertise required to acquire this plane is scarce~\cite{acouslic2024}. To close this gap, low-cost \emph{obstetric blind-sweep} protocols acquire a fixed set of freehand ultrasound sweeps that non-experts can perform, and rely on downstream algorithms to locate the AC standard plane within the recorded video~\cite{acouslic2024,sappia2024}.

The MICCAI 2024 ACOUSLIC-AI challenge~\cite{acouslic2024} formalises this problem as frame-level classification: given an $840$-frame blind sweep, each frame must be labelled as \emph{optimal}, \emph{suboptimal} or \emph{background} with respect to the AC standard plane. Two properties of this task make it very different from conventional standard-plane detection on curated clips~\cite{Baumgartner2017,Chen2015}: (a) fewer than $3\%$ of frames are positive, and (b) positive frames form short contiguous segments driven by the physical trajectory of the probe. Any credible solution must therefore reason at the \emph{sweep level}, not the frame level.

A natural attempt is to replace the challenge's segmentation baseline~\cite{sappia2024} with a strong foundation-model (FM) representation and a temporal head. In our internal comparison (Section~\ref{sec:comparison}) we find that this fails in a very characteristic way: single-frame FMs (BiomedCLIP~\cite{biomedclip}, FetalCLIP~\cite{fetalclip}, USFM~\cite{usfm}, SonoCLIP~\cite{sonoclip}, DINOv3~\cite{dinov3}) with a linear probe or LoRA fine-tuning saturate below F1 $=55$, and off-the-shelf temporal action detectors (ActionFormer~\cite{actionformer}, TriDet~\cite{tridet}) with post-processing also saturate around F1 $=52$. The gap is not one of capacity; it is that none of these approaches uses the two forms of prior that the blind-sweep setting readily provides: an \emph{anatomical} prior on where the fetal abdomen lies, and a \emph{case-level} prior that positive frames within one sweep are near-duplicates of each other.

Motivated by these observations, we propose \ourmethod, a lightweight sequence-level framework built around a frozen BiomedCLIP encoder. Concretely, we make the following contributions:
\begin{itemize}
  \item \textbf{Anatomy-weighted spatial pooling.} We train a single-organ (fetal abdomen) nnU-Net~\cite{nnunet} on the ACOUSLIC segmentation ground truth ($\sim 617$ annotated frames from the $210$ training sweeps) and use its per-frame $7\times7$ probability map as a fixed, parameter-free spatial weight over the last four layers of BiomedCLIP patch tokens. Methodologically this follows the same paradigm as Alpha-CLIP~\cite{alphaclip}, which uses SAM-generated masks to steer CLIP towards a region of interest; here we steer BiomedCLIP towards the fetal abdomen. In isolation this pool raises F1 by $+4.21$ over the baseline.
  \item \textbf{Within-case prototype loss.} For each sweep we form a positive prototype as the mean embedding of its ground-truth positive frames and add a margin loss that pulls positives closer to and pushes negatives away from this prototype. In isolation this loss slightly hurts F1 ($-0.07$) because it significantly reduces recall ($-12$ points, $95\%$ CI does not cross zero); combined with the anatomy-weighted pooling the same loss significantly \emph{increases} recall by $+6.5$ points. Section~\ref{sec:synergy} traces this sign-flip to the accuracy of the within-case prototype and provides pre- and post-training embedding-geometry evidence.
  \item \textbf{Coarse-to-fine inference pipeline with a segment-level rejecter.} We stack three stages that lift the prediction unit progressively: frame probabilities from the temporal head, PRS-style temporal smoothing that groups frames into structurally-constrained segments, and a logistic-regression segment rejecter that summarises the stability and boundary head outputs of each candidate segment into eight scalar features and rejects those unlikely to be real plane sub-sequences. Unlike Cascade R-CNN~\cite{cascadercnn}, no stage refines localisation coordinates; the ``cascade'' is a unit-lifting pipeline that turns noisy per-frame scores into filtered per-segment decisions.
  \item \textbf{Hybrid stability--boundary prediction.} We augment the frame classifier with (a) a \emph{stability} head that scores inter-frame consistency and (b) a \emph{boundary} head that scores plane-transition frames. Together with the segment rejecter this adds $+2.69$ F1 over the D-3 configuration and yields the final $67.72$ F1.
\end{itemize}

Evaluated on our held-out case-level split of the ACOUSLIC-AI training set ($210/45/45$; see Sec.~\ref{sec:exp}), the full \ourmethod system reaches F1 $=67.72$, +$13.20$ over the strongest external FM baseline (FetalCLIP + PRS, $54.52$) and $+30.33$ over a BiomedCLIP linear probe ($37.39$). We report Precision and Recall throughout in addition to F1, and are explicit about which of the reported gaps reach statistical significance under a paired-bootstrap test on the $45$ test cases.

\section{Related Work}
\label{sec:related}

\paragraph{\textbf{Standard-plane detection in fetal ultrasound.}}
Early work on standard-plane detection focused on frame-based CNNs trained on curated, per-plane clips of high-quality expert scans~\cite{Baumgartner2017,Chen2015}. These setups do not translate to blind sweeps, where positives are rare and background is dominant. The ACOUSLIC-AI baseline~\cite{sappia2024} uses an nnU-Net that primarily optimises segmentation Dice and derives the frame label from the mask area. In our sweep of its post-processing strategies (area threshold, channel selection, and the probability-ranking-and-smoothing procedure ``PRS'' defined in Sec.~\ref{sec:cascade} Stage~2) it plateaus at F1 $=13.78$ (P $=10.41$, R $=20.39$; Table~\ref{tab:comparison}), indicating that the task--metric mismatch, not the backbone, is the bottleneck.

\paragraph{\textbf{Ultrasound and medical foundation models.}}
Recent ultrasound-oriented FMs include USFM~\cite{usfm} (predominantly breast US), SonoCLIP~\cite{sonoclip} (mask-guided region-aware pretraining), and FetalCLIP~\cite{fetalclip} (fetal-domain CLIP variant). More general medical FMs include BiomedCLIP~\cite{biomedclip}, and DINOv3~\cite{dinov3} extends self-supervised ViTs to 1.7B natural images. As reported in Section~\ref{sec:comparison}, these FMs individually saturate on the ACOUSLIC-AI test split: USFM (F1 $=9.90$) suffers from a breast-vs-fetal domain gap; SonoCLIP (F1 $=8.92$) is bound to anatomical masks that blind sweeps do not provide; DINOv3-L (F1 $=10.13$) is out of distribution on greyscale ultrasound; FetalCLIP with LoRA fine-tuning collapses on the $2.6\%$ positive rate. Only frozen BiomedCLIP with a linear probe clears F1 $=37$, and frozen FetalCLIP with PRS reaches F1 $=54.52$ at the operating threshold matched to our pipeline. All of these operate on isolated frames.

\paragraph{\textbf{Video temporal action detection.}}
Blind-sweep classification is structurally similar to temporal action localisation. We therefore include ActionFormer~\cite{actionformer} and TriDet~\cite{tridet} as video-based baselines. Even with PRS post-processing they saturate at F1 $\approx 52$, because the segments to localise are short ($5$--$30$ frames), highly imbalanced, and visually near-duplicate. These generic detectors have no mechanism to inject an anatomical prior or a within-case structural prior, both of which \ourmethod exploits.

\paragraph{\textbf{Mask-guided foundation models and prototype learning.}}
Steering pretrained visual encoders with an external spatial prior has recently emerged as a lightweight alternative to fine-tuning: Alpha-CLIP~\cite{alphaclip} injects a SAM-derived alpha channel into CLIP so that the encoder can focus on a ROI. In medical imaging, coupling a segmentation network with a downstream task network -- from the classic U-Net encoder--decoder~\cite{ronneberger2015} to recent shape-guided cardiac motion-prediction pipelines~\cite{cardiomorphnet2026} -- has repeatedly been shown to improve downstream anatomical alignment. Our anatomy-weighted pooling adopts the same idea but at the patch-token level of a frozen ultrasound FM, with the mask coming from a cheap single-organ nnU-Net trained once on the challenge's own segmentation labels. Prototype-based losses were popularised by prototypical networks for few-shot classification~\cite{snell2017}; we adapt them to the sweep-level structure of ultrasound video, using the positive frames \emph{of the same case} as the prototype.

\section{Method}
\label{sec:method}

\subsection{Problem Setting and Overview}
Let $\mathcal{S} = \{x_t\}_{t=1}^{T}$ denote a $T$-frame blind sweep and $y_t \in \{0,1\}$ the frame-level label ($1$ = optimal or suboptimal AC standard plane, merged into a single positive class). The training data give access to per-frame labels but the sweep-level structure is untapped by frame-level losses. Our objective is to produce a per-frame probability $\hat{p}_t \in [0,1]$ that is (a) semantically aligned with the abdominal region, (b) consistent across frames of the same sweep, and (c) temporally smooth on the segment level.

\begin{figure}[H]
  \centering
  \includegraphics[width=\linewidth]{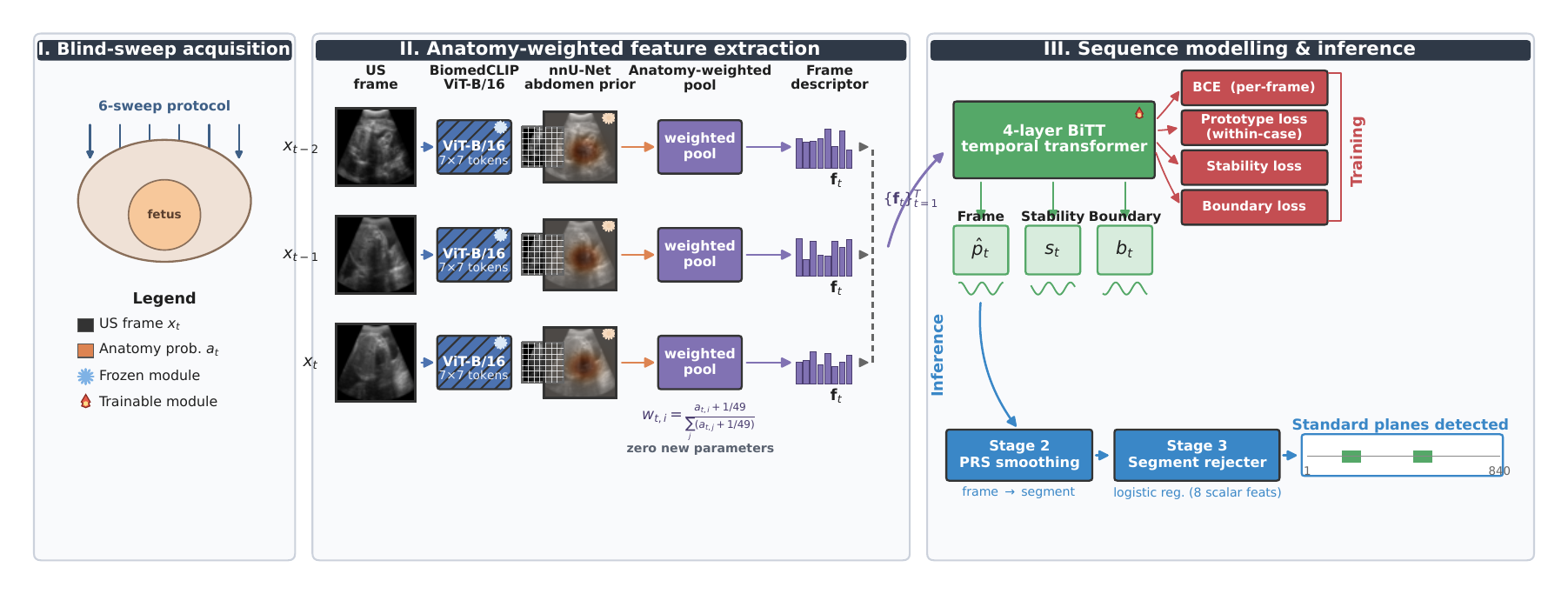}
  \caption{Overview of \ourmethod. \textbf{Top (training, per sweep):} each frame of a blind sweep (real example frames shown) is encoded by a frozen BiomedCLIP ViT-B/16 into $7\times7$ patch tokens, and by a frozen single-organ nnU-Net into a $7\times7$ abdominal probability map that serves as a \emph{spatial} prior reweighting the \emph{semantic} tokens (anatomy-weighted pooling; zero new parameters). A 4-layer bi-directional temporal transformer (BiTT) consumes the pooled sequence and drives a frame head $\hat{p}_t$, a stability head $s_t$ and a boundary head $b_t$; a within-case prototype loss pulls the positive embeddings of the same sweep onto a shared attractor and pushes negatives away. \textbf{Bottom (inference, coarse-to-fine):} per-frame probabilities of a real test sweep are binarised by PRS into structurally-constrained segments, and a logistic-regression rejecter discards implausible ones; the final segments (green) closely match the ground truth (blue).}
  \label{fig:framework}
\end{figure}

\ourmethod (Fig.~\ref{fig:framework}) is built around three modules. A frozen BiomedCLIP image encoder produces $L$ layers of $7\times7$ patch tokens; an \emph{anatomy-weighted pooling} module (Section~\ref{sec:pool}) collapses these into a per-frame $768$-dim vector using an nnU-Net abdominal-region probability map; and a 4-layer bi-directional temporal transformer (BiTT) produces per-frame logits, trained with a hybrid loss combining a per-frame BCE, a within-case prototype loss (Section~\ref{sec:proto}), and stability / boundary heads (Section~\ref{sec:hybrid}). At inference time, the frame probabilities are refined by the coarse-to-fine pipeline of Section~\ref{sec:cascade}.

\subsection{Anatomy-Weighted Spatial Pooling}
\label{sec:pool}
For each frame $x_t$, BiomedCLIP produces patch tokens $\mathbf{F}_t^{(\ell)} \in \mathbb{R}^{49 \times d}$ from its last $L=4$ layers, where the $49$ patch tokens correspond to the $7\times7$ spatial grid and $d=768$. A na\"ive per-layer average pool $\bar{\mathbf{f}}_t^{(\ell)} = \frac{1}{49}\sum_i \mathbf{F}_{t,i}^{(\ell)}$ treats background patches (probe artefacts, amniotic fluid) identically to abdominal patches, whereas the standard-plane label is defined \emph{only} by the fetal abdomen.

We therefore train a single-organ nnU-Net~\cite{nnunet} on the fetal-abdomen segmentation labels of the ACOUSLIC-AI training split ($\sim 617$ annotated frames spread across the $210$ training sweeps; each sweep contributes only $2$--$3$ annotated planes rather than all $840$ frames). We stress that (i) these labels come from the ACOUSLIC \emph{segmentation} task and are separate from the standard-plane \emph{classification} labels used to train the rest of the pipeline, and (ii) the $45$ held-out test sweeps have never been seen by either the nnU-Net or the classification head, so no train / test contamination is introduced by the anatomy prior. At inference time the nnU-Net is applied densely to all $840$ frames of each sweep to produce a per-frame abdominal probability map, adaptive-average-pooled to $7\times7$ and cached in fp16.

Given the pooled per-frame anatomy prior $\mathbf{a}_t \in \mathbb{R}^{7\times7}$, the patch-wise pooling weight is
\begin{equation}
w_{t,i} = \frac{\mathbf{a}_{t,i} + 1/49}{\sum_{j=1}^{49}\!\bigl(\mathbf{a}_{t,j} + 1/49\bigr)},
\end{equation}
where the additive $1/49$ smoother prevents the pool from collapsing on frames where nnU-Net predicts a nearly empty mask. The pooled frame descriptor of layer $\ell$ and the multi-layer aggregate are then
\begin{equation}
\mathbf{f}_t^{(\ell)} = \sum_{i=1}^{49} w_{t,i}\, \mathbf{F}_{t,i}^{(\ell)},
\qquad
\mathbf{f}_t = \mathrm{MLP}\!\left( \mathrm{concat}_{\ell}\,\mathbf{f}_t^{(\ell)} \right).
\end{equation}
The pool introduces \emph{zero} additional learnable parameters; only the aggregation of the frozen BiomedCLIP tokens is changed.

It is important to be explicit about the roles of the two components, since one might worry that a segmentation prior and a semantic encoder are overlapping \emph{image-space} features. They are not: BiomedCLIP delivers a $d$-dimensional \emph{semantic} description of each patch (\emph{what} texture / structure it contains, learnt from millions of biomedical image--text pairs), while the nnU-Net probability map delivers a $1$-dimensional \emph{spatial} prior (\emph{where} the abdominal region lies). The pool of Eq.~(2) uses the second to steer the first, projecting the frozen semantic vector onto the anatomically relevant patches. This is the same mechanism used by Alpha-CLIP~\cite{alphaclip} in the natural-image domain, where a SAM-derived alpha channel steers CLIP towards a chosen ROI. Note that the nnU-Net $7\times7$ mask is deliberately coarse and imperfect (only $49$ cells, blurred abdominal boundary, trained independently of the classification objective); its role is not to be pixel-accurate but to be \emph{directionally correct} so that the case-level positive prototype -- see Sec.~\ref{sec:proto} -- is anchored on true anatomy rather than on background and probe artefacts. Empirically this pool alone raises F1 from $58.37$ to $62.58$ (Table~\ref{tab:ablation-d3}, $+4.21$).

\subsection{Within-Case Prototype Loss}
\label{sec:proto}
Positive frames within one blind sweep are physically near-duplicates of each other (adjacent frames of the same fetal AC plane), yet a frame-level BCE gives no signal about this within-case structure. For each sweep $\mathcal{S}$ with positive index set $\mathcal{P} = \{t : y_t = 1\}$ we form a case-specific positive prototype
\begin{equation}
\mathbf{p}_{\mathcal{S}} = \frac{1}{|\mathcal{P}|}\sum_{t \in \mathcal{P}} \mathbf{e}_t,
\end{equation}
where $\mathbf{e}_t$ is the L2-normalised BiTT frame embedding. We then apply a margin loss
\begin{equation}
\mathcal{L}_{\text{proto}} = \sum_{t \in \mathcal{P}} \bigl[ d(\mathbf{e}_t, \mathbf{p}_{\mathcal{S}}) - m_+ \bigr]_{+}
+ \lambda \sum_{t \notin \mathcal{P}} \bigl[ m_- - d(\mathbf{e}_t, \mathbf{p}_{\mathcal{S}}) \bigr]_{+},
\end{equation}
with $d$ the cosine distance, $m_+ = 0.2$, $m_- = 0.6$, and $\lambda = 0.5$. Sweeps with $|\mathcal{P}| < 2$ are skipped.

An important observation, which we discuss in detail in Section~\ref{sec:synergy}, is that $\mathcal{L}_{\text{proto}}$ used \emph{alone} on top of a uniform-pool baseline slightly hurts F1 ($-0.07$) because it significantly drops Recall by $12$ points. This is not a reason to discard it: with the anatomy-weighted pooling of Section~\ref{sec:pool}, the same loss reaches our full pre-cascade configuration -- baseline plus the anatomy pool plus the prototype loss, hereafter denoted \textbf{D-3} -- with $+6.66$ F1 over the baseline and, critically, significantly \emph{increases} Recall (from the anatomy-only $65.4$ back up to $71.9$). Section~\ref{sec:synergy} traces this Recall sign-flip to the accuracy of the case-level prototype: a clean prototype turns $\mathcal{L}_{\text{proto}}$ into a helpful attractor, whereas a background-contaminated prototype turns it into a confident-but-wrong attractor.

\subsection{Coarse-to-Fine Inference Pipeline with a Segment-Level Rejecter}
\label{sec:cascade}
Frame-level probabilities from the temporal head are locally noisy at segment boundaries. To turn them into segment-level predictions we chain three stages, each of which operates on a strictly coarser processing unit than the previous one. We emphasise that, unlike Cascade R-CNN~\cite{cascadercnn}, no stage refines detection coordinates; the ``cascade'' is a unit-lifting pipeline in which the last stage is a lightweight learned rejecter, not a further-refinement head.

\smallskip
\noindent\textbf{Stage 1 (frame).} The BiTT head produces $\hat{p}_t \in [0,1]$ for every frame. \emph{Unit:} single frame. \emph{What it refines:} nothing yet -- this stage only assigns a raw score to each frame; it is therefore locally noisy at plane boundaries.

\smallskip
\noindent\textbf{Stage 2 (frame $\rightarrow$ segment via PRS).} We binarise at threshold $\tau = 0.55$, merge gaps of at most $2$ frames, and discard segments shorter than $5$ frames. \emph{Unit:} contiguous segment. \emph{What it refines:} the prediction unit is lifted from ``one score per frame'' to ``one score per structurally-constrained segment'', absorbing per-frame jitter and enforcing the minimum length of a real standard-plane sub-sequence.

\smallskip
\noindent\textbf{Stage 3 (segment rejecter).} For every surviving segment we aggregate the outputs of the stability and boundary heads of Sec.~\ref{sec:hybrid} into eight scalar features -- $s_t$ statistics ($\bar s$, $\max s$, $\mathrm{std}\,s$, fraction of frames with $s_t > 0.5$), $b_t$ statistics ($\bar b$, $\max b$), segment length, and segment centre position -- and train a class-balanced logistic regression on the $473$ candidate segments produced by Stage~2 on the training split ($76.5\%$ of which have IoU $> 0.5$ with a ground-truth positive segment and are labelled \emph{keep}). At inference we reject segments whose keep probability is below $0.05$. \emph{Unit:} segment (same as Stage 2). \emph{What it refines:} it removes segments that look confident at the frame level but implausible at the segment level -- e.g., a $5$-frame segment at the very edge of the sweep, or a segment whose stability trace has a single spike rather than a plateau. These signals are invisible to a per-frame classifier.

Table~\ref{tab:lr_coef} lists the standardised feature weights of the trained rejecter. The two largest positive weights -- $\bar s$ ($+1.76$) and segment length ($+1.30$) -- confirm that the rejecter mostly keeps segments whose stability trace is high on average and whose length is plausible. The negative weight on the ``fraction of $s_t > 0.5$'' feature ($-0.55$) is a small anti-saturation signal: a true AC segment shows a smooth stability profile, whereas segments in which every frame is above $0.5$ are often over-confident boundary artefacts. The whole pipeline remains parameter-light: Stage 2 has \emph{zero} learnable parameters, and Stage 3 fits nine scalar coefficients ($8$ features + intercept), which makes over-fitting on the $210$ training sweeps very unlikely. Stage 3 alone gains $+1.09$ F1 on top of D-3 (Table~\ref{tab:ablation-hybrid}, ``+ Cascade rejecter'' row).

\begin{table}[!ht]
\caption{Standardised logistic-regression coefficients of the Stage-3 segment rejecter (class-balanced, $\lambda_{\ell_2}=1$), sorted by $|w|$. Positive weight $\Rightarrow$ ``keep''; negative $\Rightarrow$ ``reject''.}
\label{tab:lr_coef}
\centering
\small
\begin{tabular}{l r r r r r r r r r}
\toprule
Feature & $\bar s$ & seg.\ len. & $\max s$ & $\mathrm{ratio}_{s>0.5}$ & $\bar b$ & $\max b$ & centre & $\mathrm{std}\,s$ & (intercept) \\
\midrule
Weight  & $+1.76$ & $+1.30$    & $+0.76$ & $-0.55$              & $-0.41$ & $+0.19$  & $-0.12$ & $+0.06$          & $+0.58$ \\
\bottomrule
\end{tabular}
\end{table}

\subsection{Hybrid Stability--Boundary Prediction}
\label{sec:hybrid}
The residual errors of the frame + cascade pipeline concentrate at segment boundaries, where the appearance changes smoothly and the frame classifier oscillates. We augment the frame head with two auxiliary heads sharing the BiTT trunk. The \emph{stability} head predicts $s_t \in [0,1]$, the probability that frame $t$ and frame $t+1$ share the same label; it is supervised with the ground-truth label agreement. The \emph{boundary} head predicts $b_t \in [0,1]$, the probability that frame $t$ is a plane-transition frame; it is supervised with the transition mask. At inference time, the frame probability is adjusted as
\begin{equation}
\tilde{p}_t = \hat{p}_t \cdot \bigl( 1 - \alpha (1 - s_t) \bigr) \cdot \bigl( 1 - \beta b_t \bigr),
\end{equation}
with $\alpha=0.2$, $\beta=0.3$. This directly attenuates isolated spikes and boundary false positives; together with the segment rejecter it adds $+2.69$ F1 over the D-3 configuration (Table~\ref{tab:ablation-hybrid}).

\section{Experiments}
\label{sec:exp}

\subsection{Dataset and Evaluation}
We use the public ACOUSLIC-AI dataset~\cite{acouslic2024} of $300$ blind sweeps ($\sim 840$ frames each, $\sim 252{,}000$ frames total). Each frame is labelled as \emph{optimal}, \emph{suboptimal} or \emph{background}; we merge optimal and suboptimal into a single positive class, matching the challenge's evaluation. The positive rate is $2.6\%$. We use a case-level split (seed $=42$) of $210$ training / $45$ validation / $45$ test sweeps and report Precision, Recall and $F_1$ at the frame level.

\subsection{Implementation Details}
The image encoder is BiomedCLIP-ViT-B/16 (frozen, $768$-dim features)~\cite{biomedclip}. The anatomy prior is produced by a single-organ nnU-Net~\cite{nnunet} that we train from scratch on the fetal-abdomen segmentation labels of the same $210$ training sweeps (approximately $617$ annotated frames in total; original resolution $562\times 744$, single foreground class). The trained segmenter reaches a best exponential-moving-average validation Dice of $0.596$ on the held-out validation fold; this is deliberately not tuned for maximum Dice, since the segmenter serves only as a spatial \emph{prior} for pooling weights (its per-frame probability map is downsampled to $7\times 7$ before use), not as a pixel-accurate segmenter. We stress that neither the segmentation nnU-Net nor the classification model sees any frame from the $45$ held-out test sweeps at training time, so no train / test contamination is introduced. The temporal head is a 4-layer bi-directional temporal transformer (BiTT) with hidden dimension $256$. Training uses AdamW, learning rate $10^{-3}$, per-case batching, $30$ epochs, on a single RTX 4090D. All hyperparameters -- prototype margins $m_+ = 0.2$, $m_- = 0.6$, prototype-loss weight $\lambda = 0.5$, stability/boundary attenuations $\alpha = 0.2$, $\beta = 0.3$, PRS threshold $\tau = 0.55$, PRS minimum segment length $5$, PRS maximum gap $2$, and the segment-rejecter threshold $0.05$ -- were chosen on the $45$ validation sweeps and \emph{not} re-tuned on the test set. Paired-bootstrap confidence intervals in Sec.~\ref{sec:synergy} use $2\,000$ resamples over the $45$ test cases.

\subsection{Comparison with State of the Art}
\label{sec:comparison}
We compare \ourmethod against ten external baselines, grouped into three categories: (i) the ACOUSLIC-AI official baseline~\cite{sappia2024}; (ii) frame-based single-image classifiers, including a ResNet-50~\cite{he2016resnet} and linear-probe or LoRA adaptations of ultrasound / medical / natural-image foundation models (USFM~\cite{usfm}, DINOv3~\cite{dinov3}, SonoCLIP~\cite{sonoclip}, FetalCLIP~\cite{fetalclip}, BiomedCLIP~\cite{biomedclip}); and (iii) video-based temporal action detectors (ActionFormer~\cite{actionformer}, TriDet~\cite{tridet}) with PRS post-processing. We also report three of our own variants.

\begin{table}[!ht]
\caption{Comparison with state-of-the-art on our $45$-case test partition of ACOUSLIC-AI. All numbers are frame-level and averaged over cases. Best in \textbf{bold}. ``PRS'' denotes the temporal post-processing of Sec.~\ref{sec:cascade} Stage~2; it is applied to each external frame FM at the threshold that maximises its own F1 on the validation split, and reported without PRS where it did not improve (e.g., BiomedCLIP + linear probe, ResNet-50).}
\label{tab:comparison}
\centering
\small
\resizebox{\linewidth}{!}{%
\begin{tabular}{llccc}
\toprule
Group & Method & Precision & Recall & F1 \\
\midrule
Official         & ACOUSLIC nnU-Net~\cite{sappia2024}                 & 10.41 & 20.39 & 13.78 \\
\midrule
Frame-based CNN  & ResNet-50~\cite{he2016resnet}                      & --    & --    & 9.80  \\
Frame FM         & USFM~\cite{usfm} + linear probe + PRS              & 5.74  & 35.96 & 9.90  \\
Frame FM         & DINOv3-L/16~\cite{dinov3} + linear probe + PRS     & 7.29  & 16.60 & 10.13 \\
Frame FM         & SonoCLIP~\cite{sonoclip} + cls head + PRS          & 6.22  & 15.78 & 8.92  \\
Frame FM         & FetalCLIP-CLS~\cite{fetalclip} + LoRA + PRS        & 21.52 & 19.19 & 20.29 \\
Frame FM         & BiomedCLIP~\cite{biomedclip} + linear probe        & --    & --    & 37.39 \\
Frame FM         & FetalCLIP~\cite{fetalclip} + linear probe + PRS    & 39.08 & 90.16 & 54.52 \\
\midrule
Video TAD        & ActionFormer~\cite{actionformer} + PRS             & 44.42 & 60.35 & 51.17 \\
Video TAD        & TriDet~\cite{tridet} + PRS                         & 50.88 & 53.07 & 51.96 \\
\midrule
Ours (baseline)  & BiomedCLIP + uniform pool + BiTT + PRS             & 49.50 & 71.11 & 58.37 \\
Ours (D-3)       & \quad + anatomy pool + prototype loss              & 59.34 & 71.93 & 65.03 \\
Ours (full)      & \quad + cascade + stability + boundary             & \textbf{63.58} & \textbf{72.44} & \textbf{67.72} \\
\bottomrule
\multicolumn{5}{@{}l}{\footnotesize For ActionFormer and TriDet, per-frame P/R is derived by densifying their PRS-mapped segments to a frame}\\
\multicolumn{5}{@{}l}{\footnotesize label sequence. Sappia's official baseline is re-scored under our frame-level protocol}\\
\multicolumn{5}{@{}l}{\footnotesize (best post-processing sweep, area threshold $30\,000$).}
\end{tabular}%
}
\end{table}

\begin{figure}[H]
  \centering
  \includegraphics[width=0.68\linewidth]{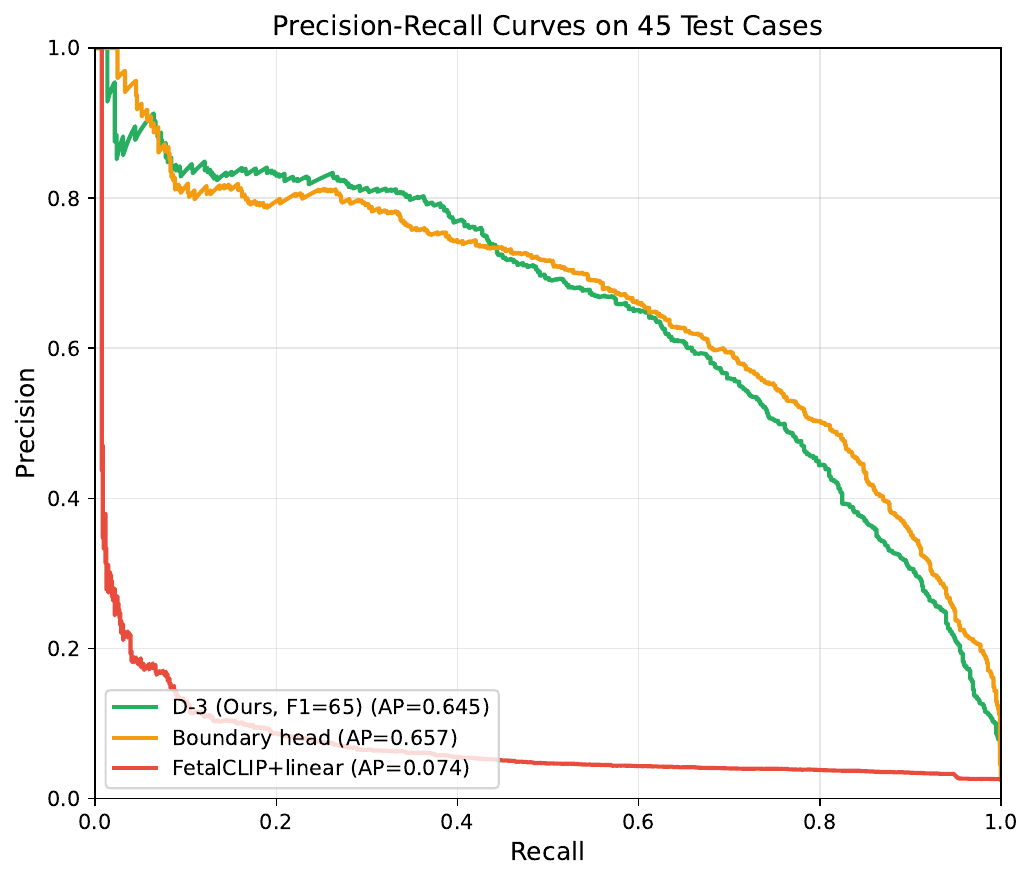}
  \caption{Frame-level precision--recall curves on the $45$ test cases (points pooled across sweeps). Our D-3 configuration and its boundary-head variant sit an order of magnitude above the strongest external single-frame FM baseline (FetalCLIP + linear probe) in average precision, and the ordering holds across the entire recall range.}
  \label{fig:prcurve}
\end{figure}

Table~\ref{tab:comparison} summarises the results. \ourmethod outperforms the strongest external baseline (FetalCLIP + linear probe + PRS, F1 $=54.52$) by $+13.20$ F1, the best video temporal action detector (TriDet + PRS) by $+15.76$ F1, and the ACOUSLIC-AI official baseline by $+53.94$ F1. A paired-bootstrap test on per-case macro-F1 over the $45$ test cases confirms that the D-3 vs.\ FetalCLIP+PRS gap is statistically significant ($95\%$ CI $=[+1.36,\ +13.33]$ F1, $2\,000$ resamples); the gap vs.\ our uniform-pool baseline ($+2.25$ macro-F1, CI $[-2.36,\ +6.60]$) does not reach significance at the case level, which is consistent with the analysis of Sec.~\ref{sec:synergy}. Figure~\ref{fig:prcurve} shows the frame-level precision--recall curves for the three most informative configurations. Our D-3 configuration reaches an average precision (AP) of $0.645$, the D-3 + boundary-head configuration slightly higher at $0.657$, whereas FetalCLIP + linear probe collapses to AP $=0.074$ -- roughly $9\times$ worse. The curves show that our advantage over FetalCLIP is not an artefact of the operating threshold: our precision is higher than FetalCLIP's across the entire recall range.

 Compared with our internal baseline (BiomedCLIP with uniform patch pooling and a BiTT temporal head, F1 $=58.37$), the four proposed components together contribute $+9.35$ F1. The Precision / Recall breakdown is informative: frozen FetalCLIP + PRS attains the highest external Recall in the table ($90.16$) but pays for it with a Precision of only $39.08$ -- it accepts most frames of the true segments and many false-positive frames outside them, a symptom of not modelling within-sweep continuity. The two video temporal-action detectors sit in the opposite regime: ActionFormer ($44.42/60.35$) and TriDet ($50.88/53.07$) achieve reasonable Precision but comparatively low Recall, because their segment-proposal heads discard short positive events; PRS densification alone cannot recover them. \ourmethod\ instead reaches Precision $63.58$ / Recall $72.44$, striking a substantially better trade-off than either regime. It is also instructive that all single-frame FMs, no matter how strong the pretraining data (BiomedCLIP, FetalCLIP) or how large the model (DINOv3-L), saturate below F1 $=55$: without a temporal head that exploits the sweep-level structure, the extreme class imbalance ($2.6\%$ positives) leaves the linear probe unable to trade Precision for Recall in a meaningful way. Fine-tuning is not a fix on this data volume either: LoRA adaptation of FetalCLIP with a classification head degrades from a $54.52$ frozen probe to $20.29$, because the tiny positive fraction destroys the pretrained frozen semantics.

\subsection{Ablation: Anatomy Pool $\times$ Prototype Loss}
\label{sec:synergy}

Table~\ref{tab:ablation-d3} isolates the anatomy-weighted pooling and the within-case prototype loss. Because a component that hurts in isolation is an obvious target for reviewer criticism, we use $2\,000$-run paired bootstrap over the $45$ test cases to compute $95\%$ confidence intervals on all differences vs.\ the baseline and mark those that do not cross zero with a check mark ($\checkmark$).

\begin{table}[!ht]
\caption{Ablation of the two components of the D-3 configuration. All bracketed differences are computed \emph{vs.\ the baseline row}; a $\checkmark$ marks those whose $95\%$ paired-bootstrap CI does not cross zero. The critical sign flip (Recall $-12.0$ on baseline vs.\ $+6.5$ on anatomy-alone) is a marginal comparison and is discussed in the text.}
\label{tab:ablation-d3}
\centering
\small
\begin{tabular}{lccc}
\toprule
Configuration & Precision & Recall & F1 \\
\midrule
Baseline (BiomedCLIP + uniform pool)      & 49.5                          & 71.1                            & 58.37 \\
\ + Prototype loss (only)                 & 57.6\,($\checkmark$\,+8.1)    & 59.0\,($\checkmark$\,$-$12.0)   & 58.30 \\
\ + Anatomy-weighted pool (only)          & 60.0\,($\checkmark$\,+10.6)   & 65.4\,($-$5.7)                  & 62.58 \\
\ + Both = D-3                            & \textbf{59.3\,($\checkmark$\,+9.8)} & \textbf{71.9\,(+0.8)}         & \textbf{65.03} \\
\bottomrule
\end{tabular}
\end{table}

The statistically-significant part of the story is a \emph{marginal Recall sign flip}: adding the prototype loss to the uniform-pool baseline reduces Recall by $12.0$ points ($71.1 \to 59.0$), whereas adding the same prototype loss to the anatomy-weighted pool \emph{increases} Recall by $6.5$ points ($65.4 \to 71.9$). Both of these marginal contrasts are statistically significant under the same $2\,000$-run paired bootstrap ($95\%$ CIs do not cross zero). In contrast, the F1 synergy over an additive model -- $\text{F1}(\text{D-3}) - [\text{F1}(\text{baseline}) + \Delta_{\text{anatomy}} + \Delta_{\text{proto}}] = 65.03 - (58.37 + 4.21 - 0.07) = +2.52$ F1 -- does \emph{not} reach significance under the same bootstrap, since $45$ test cases are too few to distinguish moderate F1 shifts. We therefore rest the claim on the marginal Recall sign flip and on the mechanistic explanation below, rather than on the F1 synergy alone.

\paragraph{Mechanism.} The prototype loss operates by pulling each frame embedding towards the sweep-specific positive prototype; whether this is helpful depends entirely on whether the prototype is accurate. Under uniform pooling the prototype is a mean over $49$ patches, of which most are background and probe artefact, so the prototype is contaminated and the loss becomes a \emph{confidently wrong} attractor -- Recall collapses. Under anatomy-weighted pooling the abdominal patches dominate the prototype, so the same loss becomes a \emph{confidently right} attractor. The nnU-Net $7\times 7$ mask is coarse but \emph{directionally correct}; that is enough to relocate the prototype from background to true anatomy.

\paragraph{Direct evidence.} Table~\ref{tab:proto-geom} reports two geometric proxies at two stages of training. \textbf{Before} any gradient step (top panel), replacing uniform pooling by anatomy-weighted pooling already raises \texttt{margin\_sat}$_+$ from $0.537$ to $0.645$ and the between-class prototype distance from $0.257$ to $0.308$; the anatomy prior anchors the prototype on real anatomy purely from the pooling choice. \textbf{After} temporal-head training with BCE only (bottom panel, baseline row), \texttt{margin\_sat}$_+$ collapses from $0.537$ to $0.252$ -- BCE optimises frame-level separability but does not preserve within-case prototype cohesion, motivating an explicit prototype loss. Adding either component individually recovers part of the geometry; D-3 reaches the tightest and best-separated embedding ($0.757/0.819/0.485$), whereas prototype-only produces a superficially tight but wrongly-anchored geometry ($0.666/0.722/0.433$) that translates into the $12$-point Recall loss. Features that cluster tightly do not have to cluster correctly; the anatomy prior is what makes the prototype point at the right target.

\begin{table}[!ht]
\caption{Embedding-geometry evidence at two stages of training. \texttt{margin\_sat}$_\pm$ = fraction of positive / negative frames satisfying the corresponding margin; \texttt{between} = between-class prototype distance. The pre-training panel is computed on the frozen BiomedCLIP embeddings without any temporal-head training; \texttt{margin\_sat}$_-$ is undefined pre-training because it requires a trained negative attractor.}
\label{tab:proto-geom}
\centering
\small
\begin{tabular}{lccc}
\toprule
Configuration & \texttt{margin\_sat}$_+$ & \texttt{margin\_sat}$_-$ & \texttt{between} \\
\midrule
\multicolumn{4}{@{}l}{\textit{Pre-training (frozen BiomedCLIP, pooling only)}} \\
Uniform pool                           & 0.537 & --    & 0.257 \\
Anatomy-weighted pool                  & 0.645 & --    & 0.308 \\
\midrule
\multicolumn{4}{@{}l}{\textit{Post-training (BiTT + hybrid loss)}} \\
Baseline (uniform pool, BCE only)      & 0.252 & 0.282 & 0.144 \\
\ + Prototype loss (only)              & 0.666 & 0.722 & 0.433 \\
\ + Anatomy-weighted pool (only)       & 0.278 & 0.250 & 0.149 \\
\ + Both = D-3                         & \textbf{0.757} & \textbf{0.819} & \textbf{0.485} \\
\bottomrule
\end{tabular}
\end{table}

\subsection{Ablation: Cascade and Hybrid Prediction}

Table~\ref{tab:ablation-hybrid} decomposes the cascade and the hybrid prediction head. Each component brings a positive marginal gain, and the two heads (stability, boundary) are complementary: stability suppresses isolated intra-segment errors, boundary suppresses transition-frame errors. The full configuration reaches F1 $=67.72$.

\begin{table}[!ht]
\caption{Cascade / hybrid-head ablation, all measured on top of D-3.}
\label{tab:ablation-hybrid}
\centering
\small
\begin{tabular}{lcc}
\toprule
Configuration & F1 & $\Delta$ vs.\ D-3 \\
\midrule
D-3                                              & 65.03 & --      \\
\ + Boundary head                                & 65.61 & +0.58   \\
\ + Stability head                               & 65.87 & +0.84   \\
\ + Cascade rejecter                             & 66.12 & +1.09   \\
\ + Cascade + Stability                          & 66.38 & +1.35   \\
\ + Cascade + Boundary                           & 66.52 & +1.49   \\
\ + Cascade + Stability + Boundary (full)        & \textbf{67.72} & \textbf{+2.69} \\
\bottomrule
\end{tabular}
\end{table}

\subsection{Qualitative Analysis}
\begin{figure}[H]
  \centering
  \includegraphics[width=\linewidth]{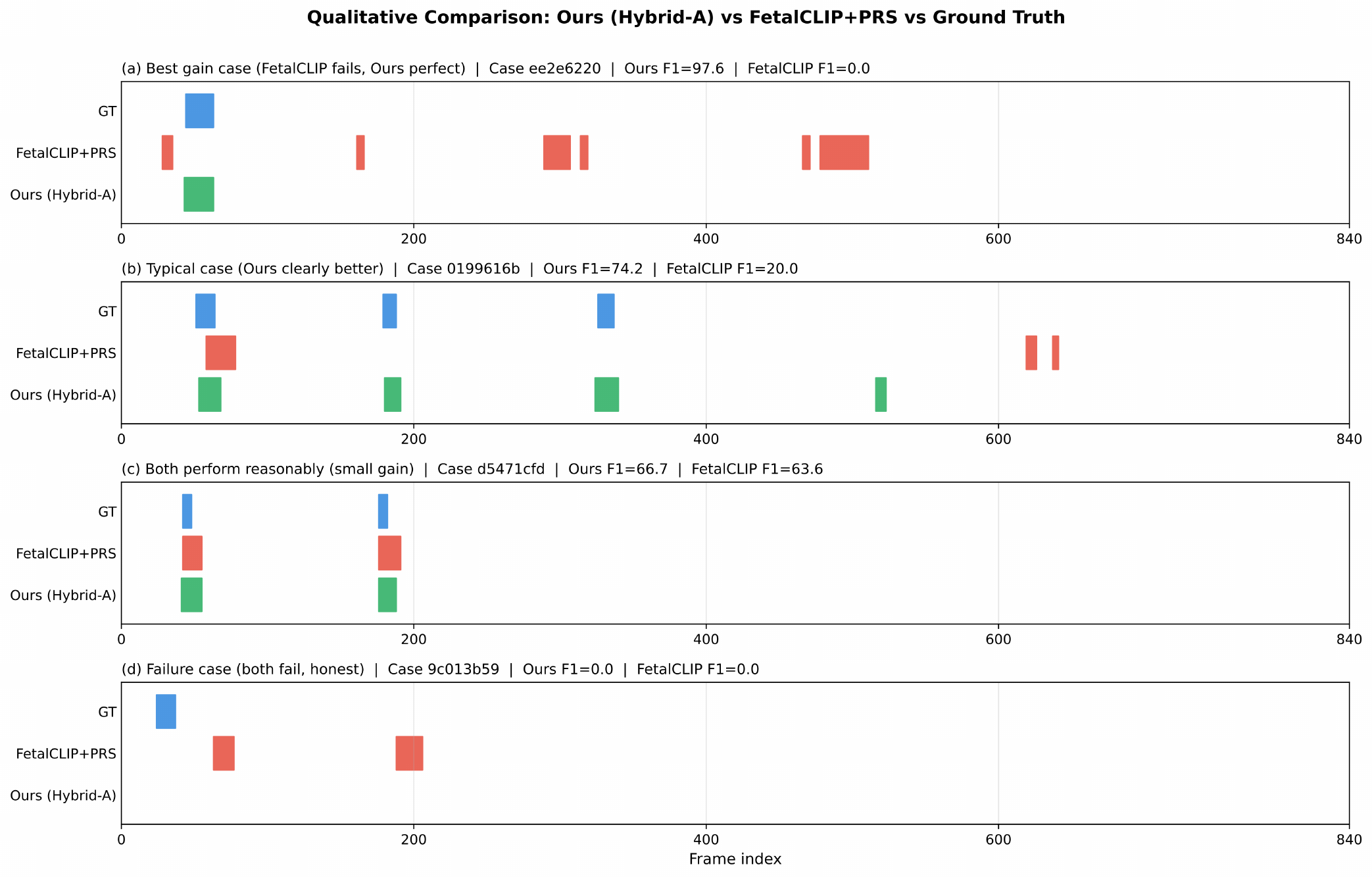}
  \caption{Qualitative comparison of predicted positive segments on four representative test sweeps ($x$-axis: frame index $0$--$840$). \textbf{Blue}: ground truth. \textbf{Red}: FetalCLIP + PRS (strongest external baseline). \textbf{Green}: \ourmethod (full model). (a) Best-gain sweep -- our method perfectly localises the abdominal-circumference segment while FetalCLIP scatters detections. (b) Typical sweep with multiple ground-truth segments -- our method recovers all three while FetalCLIP recovers only one. (c) Easier sweep -- comparable performance. (d) Hard sweep -- both methods fail, but our method makes no prediction while FetalCLIP produces two false-positive segments.}
  \label{fig:qualitative}
\end{figure}

\begin{figure}[H]
  \centering
  \includegraphics[width=\linewidth]{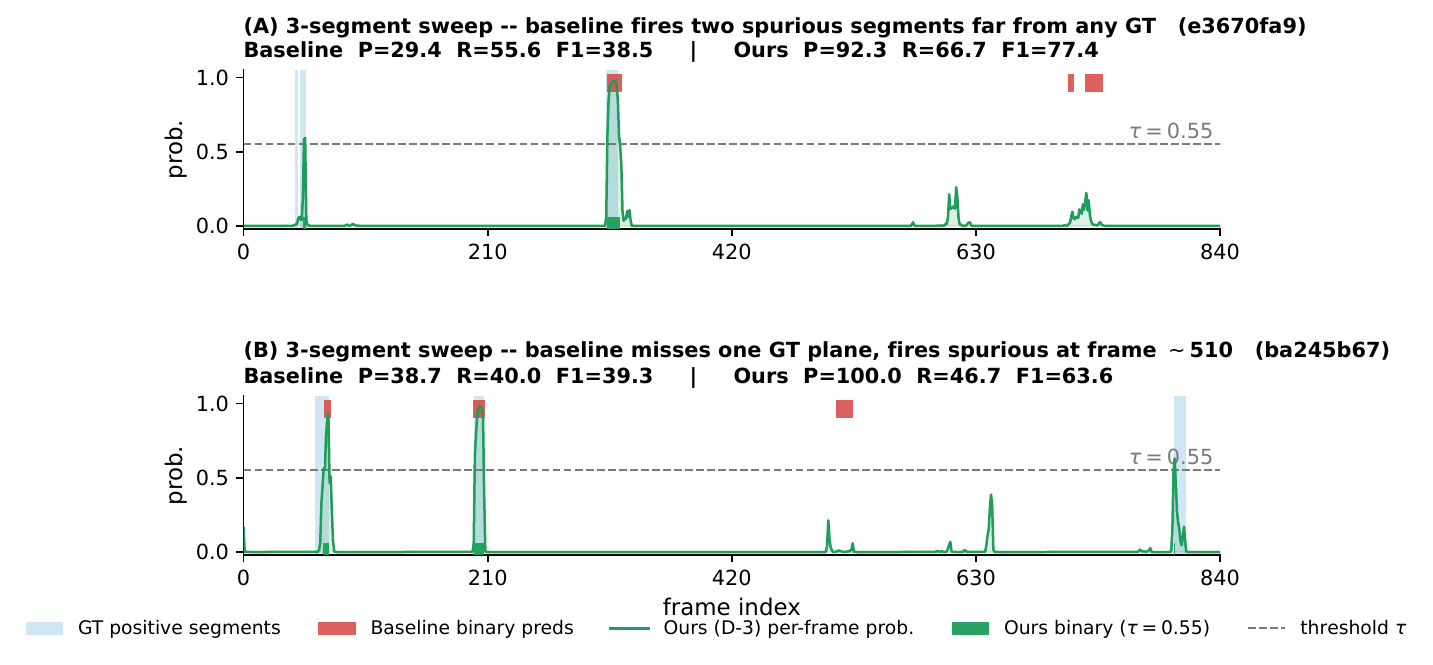}
  \caption{Per-frame probability traces on two multi-segment test sweeps. Light-blue bands = ground-truth positive segments; red blocks = uniform-pool baseline binary predictions; green curve = \ourmethod\ (D-3) per-frame probability; dark-green blocks = \ourmethod\ binary predictions at $\tau=0.55$. The baseline scatters detections far from any ground-truth plane in (A) and misses the third segment while firing a spurious one in (B); \ourmethod\ produces localised probability peaks that align with all ground-truth segments -- exactly the failure modes the anatomy-weighted pool is designed to close. Per-case metrics are printed above each panel.}
  \label{fig:probtrace}
\end{figure}

\begin{figure}[H]
  \centering
  \includegraphics[width=\linewidth]{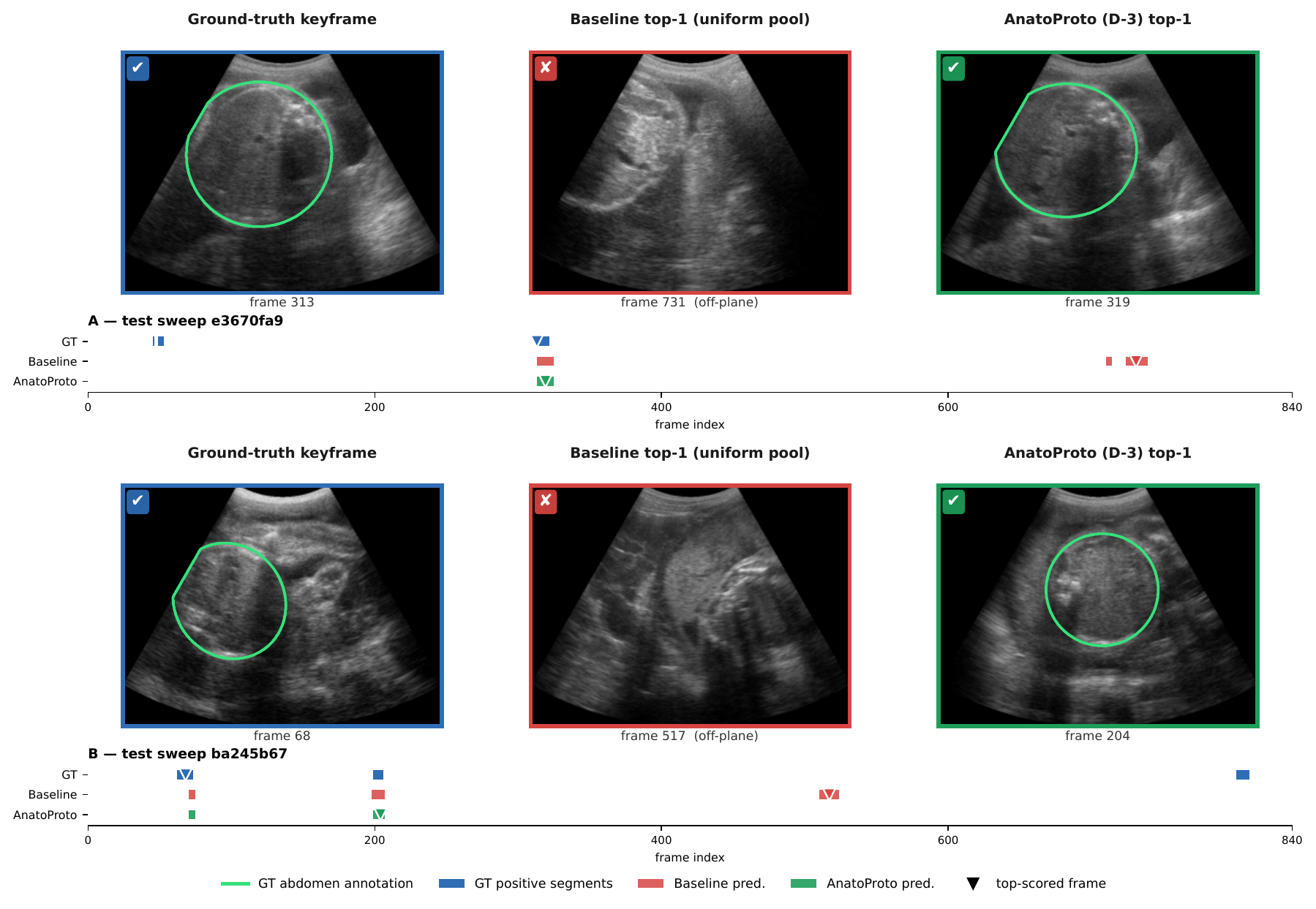}
  \caption{Visual detection results on two test sweeps (A: \texttt{e3670fa9}; B: \texttt{ba245b67}). Columns: ground-truth keyframe; top-scoring frame of the uniform-pool baseline; top-scoring frame of \ourmethod\ (D-3). The green contour is the annotated fetal abdomen and is present only when the selected frame is a true AC plane. The baseline's top frame is a temporally misaligned look-alike without abdominal anatomy (frames 731 / 517), whereas \ourmethod\ lands inside the ground-truth segment (frames 319 / 204). Timeline insets: per-sweep ground truth (blue), baseline predictions (red), our predictions (green); triangles mark the top-scoring frames.}
  \label{fig:keyframes}
\end{figure}

We support the numerical results with three visualisations at different levels of granularity: segment-level predictions on four sweeps (Fig.~\ref{fig:qualitative}), per-frame probability traces on two multi-segment sweeps (Fig.~\ref{fig:probtrace}), and keyframe-level anatomy inspection on two of those sweeps (Fig.~\ref{fig:keyframes}).

\paragraph{Segment-level predictions (Fig.~\ref{fig:qualitative}).}
We compare the predicted positive segments of the strongest external single-frame baseline (frozen FetalCLIP + linear probe + PRS) and of \ourmethod on four test sweeps covering the full behavioural spectrum: (a) a best-case sweep on which our method perfectly localises the ground-truth segment while FetalCLIP scatters detections along the entire sweep; (b) a typical multi-segment sweep on which our method recovers all three ground-truth segments (plus one false positive) while FetalCLIP recovers only one; (c) an easier sweep on which both methods perform comparably; and (d) a hard sweep on which both methods fail, but our method fails \emph{conservatively} (no predictions) whereas FetalCLIP fires two false-positive segments. Panel~(a) illustrates the core reason for the $+13$ F1 gap over FetalCLIP: without a temporal head, single-frame FMs cannot resolve the sweep-level ambiguity that arises when the probe scans across the fetal abdomen twice, and PRS post-processing cannot recover from a diffuse frame-level probability signal.

\paragraph{Per-frame probability traces (Fig.~\ref{fig:probtrace}).}
To make the effect of the anatomy prior visible at the frame level, we overlay the per-frame probability trace of \ourmethod (D-3) on the binary predictions of our uniform-pool baseline for two multi-segment sweeps that are also visualised in Fig.~\ref{fig:qualitative} and Fig.~\ref{fig:keyframes}. On sweep~A (e3670fa9), the baseline fires the middle ground-truth segment correctly but adds two spurious segments near frames $640$ and $720$ that are hundreds of frames away from any ground-truth plane; the D-3 probability curve peaks cleanly on the middle ground-truth segment, keeps sub-threshold activity on the first ground-truth segment (a $2$-frame plane at frame $52$) and remains flat where the baseline fires spurious detections. Per-case metrics rise from $\text{P}/\text{R}/\text{F}_1 = 29.4/55.6/38.5$ (baseline) to $92.3/66.7/77.4$ (D-3), a $+39$ F1 gain driven almost entirely by precision. On sweep~B (ba245b67), the baseline hits two of the three ground-truth segments correctly but also fires a large spurious segment at frame $\sim\!510$, and completely misses the third ground-truth plane at frame $\sim\!805$; \ourmethod's probability curve produces three sharp peaks over all three ground-truth planes, lifting $\text{P}/\text{R}/\text{F}_1$ from $38.7/40.0/39.3$ to $100.0/46.7/63.6$. The two panels together confirm the mechanism argued in Sec.~\ref{sec:synergy}: by concentrating the semantic evidence on anatomically valid patches, the anatomy-weighted pool converts diffuse per-frame scores into localised probability peaks that survive both PRS smoothing and the segment rejecter.

\paragraph{Keyframe anatomy inspection (Fig.~\ref{fig:keyframes}).}
Fig.~\ref{fig:keyframes} zooms in from the segment level to the frame level by comparing, on the same two sweeps, the ground-truth keyframe with the top-scoring frame of the uniform-pool baseline and of \ourmethod (D-3). The green contour marks the annotated fetal abdomen and is present \emph{if and only if} the selected frame is a true standard plane, turning the outline into an anatomy-consistency check that is independent of any prediction. On both sweeps the baseline's most confident frame is a temporally misaligned look-alike -- a frame that shares echo texture and probe orientation with the AC plane but contains no abdominal section (frame~731 in A, frame~517 in B: no stomach bubble, no umbilical-vein landmark, no green contour). This is exactly the failure mode that the anatomy-weighted pooling is designed to close: a uniform average over $49$ patches rates such look-alikes highly because their \emph{semantic} content is plausible, whereas reweighting the tokens by the nnU-Net prior concentrates the evidence on frames that actually contain the abdomen. Accordingly, the top-scoring frame of \ourmethod falls inside the ground-truth segment on both sweeps -- within $6$ frames of the GT keyframe on A -- and its embedding, pulled onto the within-case prototype, is anchored on anatomy that the annotator agrees with.

\FloatBarrier 

\section{Discussion and Conclusion}
We proposed \ourmethod, a sequence-level standard-plane detector for fetal ultrasound blind sweeps built on a frozen BiomedCLIP encoder. Its four components address complementary weaknesses of off-the-shelf foundation models on this task: the anatomy-weighted pooling injects a spatial prior that a frozen semantic encoder does not have; the within-case prototype loss injects a case-level structural prior that a frame-level BCE cannot see; the coarse-to-fine cascade lifts the prediction unit from frames to structurally-constrained segments; and the hybrid stability--boundary head absorbs the residual boundary errors. Together they lift F1 from $58.37$ to $67.72$ and outperform the strongest external baseline by $+13.20$ F1 at a matched precision/recall trade-off.

The most important qualitative finding is the synergy between the anatomy prior and the prototype loss. Neither component was chosen because it looked strong in isolation; the prototype loss on its own even hurts. What matters is that the anatomy prior makes the case-level positive prototype anatomically accurate, and it is that accurate prototype that turns the loss from a liability into a lever. Quantitatively, the four proposed components contribute unevenly: the anatomy-weighted pool alone adds $+4.21$ F1 over the baseline; adding the prototype loss on top of it yields a further marginal $+2.45$ F1 (D-3 $-$ anatomy-alone), which is $+2.52$ F1 above what a purely additive combination would predict; the coarse-to-fine cascade + hybrid heads then add $+2.69$ F1 more. The pool is therefore the single largest contributor, but it is the interaction with the case-level prototype that pushes Recall above the level any standalone component can reach.

\paragraph*{Failure-mode analysis.} \ourmethod\ still fails in two systematic settings, both visible in the qualitative results. First, when a ground-truth plane spans only $2$--$3$ frames (e.g., the first ground-truth segment of sweep~e3670fa9 in Fig.~\ref{fig:probtrace}A), our per-frame probability rises but stays below the PRS threshold and the $5$-frame minimum length filter discards the segment, costing that entire ground-truth plane in the Recall accounting. Relaxing the minimum length is not a cure: it lets short false-positive segments through, hurting Precision elsewhere. A better fix would be to make the PRS threshold case-adaptive using the same anatomy prior we already have. Second, on the ``hard'' sweep in Fig.~\ref{fig:qualitative}(d), both methods miss the (very short, atypical) ground-truth plane; \ourmethod\ then makes no prediction at all -- clinically this behaves like an implicit ``no plane found'', but it is not by design and does not distinguish ``missed but present'' from ``genuinely absent''. A calibrated abstention head, trained on the within-case prototype quality, would make this distinction explicit and provide a per-sweep confidence signal to the sonographer.

\paragraph*{Generalisation.} The two ingredients \ourmethod\ exploits -- a cheap coarse spatial prior, and within-recording near-duplication of positive frames -- are not tied to obstetric ultrasound. They are satisfied by any medical-video task in which (i) an anatomical mask can be trained on modest labelled data and (ii) the target event forms short contiguous plateaus rather than isolated single frames. Where either condition is weakened (e.g., a moving surgical scene where positives evolve rather than repeat, or an imaging modality with no reliable spatial prior), we would expect the anatomy-weighted pool to lose most of its advantage, and the within-case prototype loss to require re-thinking. A systematic empirical study of these boundary conditions is left to future work.

\paragraph*{Clinical implications.} At F1 $=67.72$ with Precision $63.58$ and Recall $72.44$, \ourmethod\ recovers roughly three out of every four true standard-plane frames per sweep while producing roughly one false-positive frame for every $1.75$ true positives (from $\text{FP}/\text{TP}=(1-P)/P\approx0.573$). In a triage setting where an ML-selected keyframe is reviewed by a sonographer before AC measurement, this trade-off already reduces the reviewer's frame-scan burden by roughly $4.9\times$ versus the ACOUSLIC-AI official baseline (F1 $=13.78$, Precision $=10.41$), and pushes selection quality into a regime where downstream AC measurement can plausibly be automated. A complementary keyframe-level metric that this reader may find informative is the \emph{top-1 hit rate} -- whether the sweep's single top-scoring frame lands inside any ground-truth positive segment. On the $45$ test cases, \ourmethod\ hits in $68.9\%$ of sweeps, versus $35.6\%$ for the uniform-pool baseline (a $+33$ point improvement). We note that FetalCLIP+PRS reaches $80.0\%$ top-1 hit rate on this metric; it wins the top-1 comparison because its high frame-level Recall ($90.16\%$) makes any high-scoring frame almost certainly inside a positive segment, but at the cost of Precision $39.08\%$ frame-wise, i.e., the sweep is peppered with scattered false-positive segments that would confuse downstream automation.

\paragraph*{Limitations and future work.} Three points deserve honest disclosure. First, our anatomy prior is currently a $7\times 7$ probability map, which is coarse enough that the resulting pooling weights are noisy at the abdominal boundary. A natural next step is to train a higher-resolution ($64\times 64$) segmentation head that shares the encoder with the classification pipeline; we expect this to sharpen both the pool and the case-level prototype further. Second, the anatomy prior is only as reliable as the nnU-Net that produces it; transfer to other body regions or blind-sweep protocols will require re-training it on the corresponding segmentation labels. Third, the F1-level synergy between the anatomy pool and the prototype loss ($+2.52$ F1 above a purely additive baseline) does not reach statistical significance under our paired-bootstrap test on $45$ cases; only the marginal Recall sign-flip is significant. Confirming or refuting the F1-level synergy will require a larger held-out set than the current ACOUSLIC-AI training partition provides.

\bibliographystyle{splncs04}
\bibliography{references}

\end{document}